\documentclass[twoside]{article}

\usepackage[accepted]{aistats2021}
\usepackage{hyperref}       
\usepackage{url}            
\usepackage{booktabs}       
\usepackage{amsfonts}       
\usepackage{amsmath}
\usepackage{nicefrac}       
\usepackage{microtype}      

\usepackage[ruled,vlined]{algorithm2e}
\usepackage{bm}
\usepackage{comment}
\usepackage{tikz}
\def\checkmark{\tikz\fill[scale=0.4](0,.35) -- (.25,0) -- (1,.7) -- (.25,.15) -- cycle;}

\usepackage{pgfplots}
\usepackage{graphicx}
\usepackage{multirow}

\begin{document}

%

%

\twocolumn[

\aistatstitle{Leveraging the Feature Distribution\\in Transfer-based Few-Shot Learning}

\aistatsauthor{ Yuqing Hu \And Vincent Gripon \And  St\'ephane Pateux }

\aistatsaddress{ IMT Atlantique \\ Orange Labs \And  IMT Atlantique \And Orange Labs } ]

\begin{abstract}
  Few-shot classification is a challenging problem due to the uncertainty caused by using few labelled samples. In the past few years, many methods have been proposed to solve few-shot classification, among which transfer-based methods have proved to achieve the best performance. Following this vein, in this paper we propose a novel transfer-based method that builds on two successive steps: 1) preprocessing the feature vectors so that they become closer to Gaussian-like distributions, and 2) leveraging this preprocessing using an optimal-transport inspired algorithm (in the case of transductive settings). Using standardized vision benchmarks, we prove the ability of the proposed methodology to achieve state-of-the-art accuracy with various datasets, backbone architectures and few-shot settings. The code can be found at~\url{https://github.com/yhu01/PT-MAP}.
\end{abstract}

\section{Introduction}
\label{introduction}

Thanks to their outstanding performance, Deep Learning methods are widely considered for vision tasks such as object classification or detection. To reach top performance, these systems are typically trained using very large labelled datasets that are representative enough of the inputs to be processed afterwards.

However, in many applications, it is costly to acquire or to annotate data, resulting in the impossibility to create such large labelled datasets. In this context, it is challenging to optimize Deep Learning architectures considering the fact they typically are made of way more parameters than the dataset contains. This is why in the past few years, few-shot learning (i.e. the problem of learning with few labelled examples) has become a trending research subject in the field. In more details, there are two settings that authors often consider: a) ``inductive few-shot'', where only a few labelled samples are available during training and prediction is performed on each test input independently, and b) ``transductive few-shot'', where prediction is performed on a batch of (non-labelled) test inputs, allowing to take into account their joint distribution.

Many works in the domain are built based on a ``learning to learn'' guidance, where the pipeline is to train an optimizer~\cite{finn2017model, ravi2016optimization,thrun2012learning} with different tasks of limited data so that the model is able to learn generic experience for novel tasks. Namely, the model learns a set of initialization parameters that are in an advantageous position for the model to adapt to a new (small) dataset. Recently, the trend evolved towards using well-thought-out transfer architectures (called backbones)~\cite{torrey2010transfer,das2019two} trained one time on the same training data, but seen as a unique large dataset.

A main problem of using feature vectors extracted using a backbone architecture is that their distribution is likely to be complex, as the problem the backbone has been optimized for most of the time differs from the considered task. As such, methods that rely on strong assumptions about the data distributions are likely to fail in leveraging the quality of features. In this paper, we tackle the problem of transfer-based few-shot learning with a twofold strategy: 1) preprocessing the data extracted from the backbone so that it fits a particular distribution (i.e. Gaussian-like) and 2) leveraging this specific distribution thanks to a well-thought proposed algorithm based on maximum a posteriori and optimal transport (only in the case of transductive few-shot). Using standardized benchmarks in the field, we demonstrate the ability of the proposed method to obtain state-of-the-art accuracy, for various problems and backbone architectures in some inductive settings and most transductive ones.

\usetikzlibrary{shapes,arrows}
\tikzstyle{block} = [draw, fill=blue!20, rectangle, 
    minimum height=3em, minimum width=6em]
\tikzstyle{sum} = [draw, fill=blue!20, circle, node distance=1cm]
\tikzstyle{input} = [coordinate]
\tikzstyle{output} = [coordinate]
\tikzstyle{pinstyle} = [pin edge={to-,dashed, thin,black}]

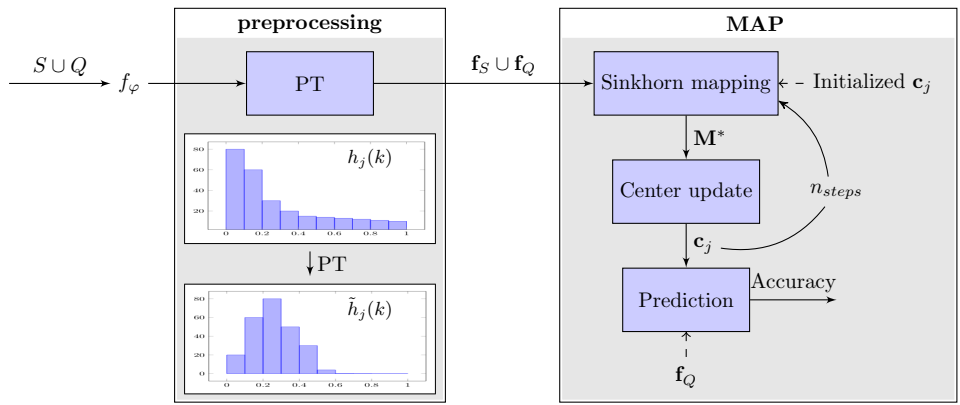
\begin{figure*}
    \centering
    \begin{tikzpicture}[auto, node distance=2cm,>=latex', scale=0.8, every node/.style={scale=0.8}]
    \draw[fill=black,fill opacity=0.1,draw=black,draw opacity=0.1]
    (-0.2,0) rectangle (4.2,6)
    (6.2,0) rectangle (12.7,6);
    \draw[]
    (-0.25,-0.05) rectangle (4.25,6.5)
    (6.15,-0.05) rectangle (12.75,6.5);
    \node at (2,6.25) {\textbf{preprocessing}};
    \node at (9.4,6.25) {\textbf{MAP}};
    
    \node [input, name=input] at (-3, 5.25){};
    \node [right of=input] (extractor) {$f_\varphi$};
    \node [block, right of=extractor,node distance=3cm] (transform) {PT};
    \draw [draw,->] (input) -- node {$S\cup Q$} (extractor);
    \draw [->] (extractor) -- node {} (transform);
    
    \node [block, right of=transform, node distance=6.25cm, pin={[pinstyle]right:Initialized $\mathbf{c}_j$}] (sinkhorn) {Sinkhorn mapping};
    \node [block, below of=sinkhorn, node distance=1.8cm] (center) {Center update};
    \node [right of=center, node distance=2.5cm] (step) {$n_{steps}$};
    \node [block, below of=center, node distance=1.8cm, pin={[pinstyle]below:$\mathbf{f}_Q$}] (prediction) {Prediction};
    \node [output, right of=prediction, node distance=2.5cm] (output) {};
     
    \draw [->] (transform) -- node[pos=0.59, name=f] {$\mathbf{f}_S\cup\mathbf{f}_Q$} (sinkhorn);
    \draw [->] (sinkhorn) -- node[name=m] {$\mathbf{M^*}$} (center);
    \draw [->] (center) -- node[name=c] {$\mathbf{c}_j$} (prediction);
    \draw [->] (prediction) -- node[name=o] {Accuracy} (output);
    \path
    (c) edge[bend right] (step);
    \path[->,>=stealth']
    (step) edge[bend right=25] (sinkhorn);
    
    \node at (2,3.5) (skew) {\fcolorbox{black}{white}{\includegraphics[width=0.23\linewidth,height=1.6cm]{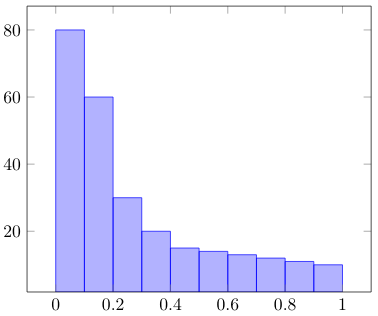}}};
    \node[font=\small] at (3,4) {$h_j(k)$};
    \node at (2,1) (gaussian) {\fcolorbox{black}{white}{\includegraphics[width=0.23\linewidth,height=1.6cm]{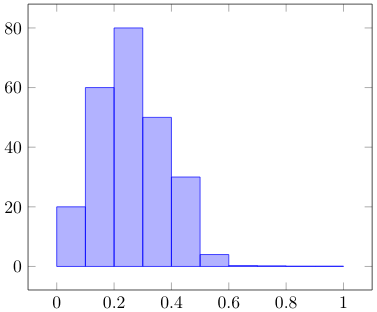}}};
    \node[font=\small] at (3,1.5) {$\tilde{h}_j(k)$};
    \draw [->] (skew) -- node[name=pt] {PT} (gaussian);
    
    \end{tikzpicture}
    \caption{Illustration of the proposed method. First we extract feature vectors of all the inputs in $\mathbf{D}_{novel}$ and preprocess them to obtain $\mathbf{f}_S\cup\mathbf{f}_Q$. Note that the Power transform (PT) has the effect of mapping a skewed feature distribution into a gaussian-like distribution ($h_j(k)$ denotes the histogram of feature $k$ in class $j$). In MAP, we perform Sinkhorn mapping with class center $\mathbf{c}_j$ initialized on $\mathbf{f}_S$ to obtain the class allocation matrix $\mathbf{M^*} $ for $\mathbf{f}_Q$, and we update the class centers for the next iteration. After $n_{steps}$ we evaluate the accuracy on $\mathbf{f}_Q$.}
    \label{fig:illustration}
\end{figure*}

\section{Related work}
\label{related work}

A large volume of works in few-shot classification is based on meta learning~\cite{thrun2012learning} methods, where the training data is transformed into few-shot learning episodes to better fit in the context of few examples. In this branch, optimization based methods~\cite{thrun2012learning, finn2017model, ravi2016optimization} train a well-initialized optimizer so that it quickly adapts to unseen classes with a few epochs of training. Other works~\cite{zhang2019few,chen2019image} utilize data augmentation techniques to artificially increase the size of the training datasets.

In the past few years, there have been a growing interest in transfer-based methods. The main idea consists in training feature extractors able to efficiently segregate novel classes it never saw before. For example, in~\cite{chen2019closer} the authors train the backbone with a distance-based classifier~\cite{mensink2012metric} that takes into account the inter-class distance. In~\cite{mangla2020charting}, the authors utilize self-supervised learning techniques~\cite{chapelle2009semi}  to co-train an extra rotation classifier for the output features, improving the accuracy in few-shot settings. Many approaches are built on top of a feature extractor. For instance, in~\cite{wang2019simpleshot} the authors implement a nearest class mean classifier to associate an input with a class whose centroid is the closest in terms of the $\ell_2$ distance. In~\cite{lichtenstein2020tafssl} an iterative approach is used to adjust the class centers. In~\cite{hu2020exploiting} the authors build a graph neural neural network to gather the feature information from similar samples. Transfer-based techniques typically reach the best performance on standardized benchmarks.

Although many works involve feature extraction, few have explored the features in terms of their distribution~\cite{8462273}. Often, assumptions are made that the features in a class align to a certain distribution, even though these assumptions are rarely experimentally discussed. In our work, we analyze the impact of the features distributions and how they can be transformed for better processing and accuracy. We also introduce a new algorithm to improve the quality of the association between input features and corresponding classes in typical few-shot settings.

\paragraph{Contributions.} Let us highlight the main contributions of this work. (1) We propose to preprocess the raw extracted features in order to make them more aligned with Gaussian assumptions. Namely we introduce transforms of the features so that they become less skewed. (2) We use a wasserstein-based method to better align the distribution of features with that of the considered classes. (3) We show that the proposed method can bring large increase in accuracy with a variety of feature extractors and datasets, leading to state-of-the-art results in the considered benchmarks.

\section{Methodology}
\label{methodology}

In this section we introduce the problem settings. We discuss the training of the feature extractors, the preprocessing steps that we apply on the trained features and the final classification algorithm. A summary of our proposed method is depicted in Figure~\ref{fig:illustration}.

\subsection{Problem statement}

We consider a typical few-shot learning problem. We are given a \emph{base} dataset $\mathbf{D}_{base}$ and a \emph{novel} dataset $\mathbf{D}_{novel}$ such that $\mathbf{D}_{base}\cap\mathbf{D}_{novel}=\emptyset$. $\mathbf{D}_{base}$ contains a large number of labelled examples from $K$ different classes. $\mathbf{D}_{novel}$, also referred to as a task in other works, contains a small number of labelled examples (support set $S$), along with some unlabelled ones (query set $Q$), all from $w$ \emph{new} classes. Our goal is to predict the class of the unlabelled examples in the query set. The following parameters are of particular importance to define such a few-shot problem: the number of classes in the novel dataset $w$ (called $w$-way), the number of labelled samples per class $s$ (called $s$-shot) and the number of unlabelled samples per class $q$. So the novel dataset contains a total of $w(s+q)$ samples, $ws$ of them being labelled, and $wq$ of them being those to classify. In the case of inductive few-shot, the prediction is performed independently on each one of the $wq$ samples. In the case of transductive few-shot~\cite{liu2018learning, lichtenstein2020tafssl}, the prediction is performed considering all $wq$ samples together. In the latter case, most works exploit the information that there are exactly $q$ samples in each class. We discuss this point in the experiments.

\subsection{Feature extraction}

The first step is to train a neural network backbone model using only the base dataset. In this work we consider multiple backbones, with various training procedures. 
Once the considered backbone is trained, we obtain robust embeddings that should generalize well to novel classes. We denote by $f_\varphi$ the backbone function, obtained by extracting the output of the penultimate layer from the considered architecture, with $\varphi$ being the trained architecture parameters.
Note that importantly, in all backbone architectures used in the experiments of this work, the penultimate layers are obtained by applying a ReLU function, so that all feature components coming out of $f_\varphi$ are nonnegative.

\subsection{Feature preprocessing}

As mentioned in Section~\ref{related work}, many works hypothesize, explicitly or not, that the features from the same class are aligned with a specific distribution (often Gaussian-like). But this aspect is rarely experimentally verified. In fact, it is very likely that features obtained using the backbone architecture are not Gaussian. Indeed, usually the features are obtained after applying a relu function, and exhibit a positive distribution mostly concentrated around 0 (see details in the next section).

Multiple works in the domain~\cite{wang2019simpleshot, lichtenstein2020tafssl} discuss the different statistical methods (e.g. normalization) to better fit the features into a model. Although these methods may have provable assets for some distributions, they could worsen the process if applied to an unexpected input distribution. This is why we propose to preprocess the obtained feature vectors so that they better align with typical distribution assumptions in the field. Namely, we use a power transform as follows.

\paragraph{Power transform (PT).} Denote $\mathbf{v} = f_\varphi(\mathbf{x})\in{\left({\mathbb{R}^+}\right)}^d, \mathbf{x}\in \mathbf{D}_{novel}$ as the obtained features on $\mathbf{D}_{novel}$. We hereby perform a power transformation method, which is similar to Tukey's Transformation Ladder~\cite{tukey1977exploratory}, on the features. We then follow a unit variance projection, the formula is given by:
\begin{equation}
f(\mathbf{v}) = \left\{\begin{array}{ll}\frac{(\mathbf{v}+\epsilon)^{\beta}}{\|(\mathbf{v}+\epsilon)^{\beta}\|_2}& \text{if } \beta \neq 0\\ \frac{\log{(\mathbf{v}+\epsilon)}}{\|\log{(\mathbf{v}+\epsilon)\|_2}}& \text{if } \beta = 0\end{array}\right.,
\label{eq:preprocessing}
\end{equation}
where $\epsilon = 1e-6$ is used to make sure that $\mathbf{v}+\epsilon$ is strictly positive and $\beta$ is a hyper-parameter. The rationales of the preprocessing above are: (1) Power transforms have the functionality of reducing the skew of a distribution, adjusted by $\beta$, (2) Unit variance projection scales the features to the same area so that large variance features do not predominate the others. This preprocessing step is often able to map data from any distribution to a close-to-Gaussian distribution. We will analyse this ability and the effect of power transform in more details in Section~\ref{experiments}.

Note that $\beta=1$ leads to almost no effect. More generally, the skew of the obtained distribution changes when $\beta$ varies. For instance, if a raw distribution is right-skewed, decreasing $\beta$ phases out the right skew, and phases into a left-skewed distribution when $\beta$ becomes negative. After experiments, we found that $\beta=0.5$ gives the most consistent results for our considered experiments. More details based on our considered experiments are available in Section~\ref{experiments}.

This first step of feature preprocessing can be performed in both inductive and transductive settings.

\subsection{MAP}

Let us assume that the preprocessed feature distribution for each class is Gaussian or Gaussian-like. As such, a well-positioned class center is crucial to a good prediction. In this section we discuss how to best estimate the class centers when the number of samples is very limited and classes are only partially labelled. In more details, we propose an Expectation–Maximization~\cite{dempster1977maximum}-like algorithm that will iteratively find the Maximum A Posteriori (MAP) estimates of the class centers. 

We firstly show that estimating these centers through MAP is similar to the minimization of Wasserstein distance. Then, an iterative procedure based on a Wasserstein distance estimation, using the sinkhorn algorithm~\cite{cuturi2013sinkhorn,vallender1974calculation,huang2019few}, is designed to estimate the optimal transport from the initial distribution of the feature vectors to one that would correspond to the draw of samples from Gaussian distributions. 

Note that in this step we consider what is called the ``transductive'' setting in many other few shot learning works~\cite{liu2018learning, lichtenstein2020tafssl, liu2019prototype, hu2020exploiting, kye2020transductive, garcia2017few, kim2019edge, gidaris2019generating, ye2018learning}, where we exploit unlabelled samples during the procedure as well as priors about their relative proportions. 

In the following, we denote by $\mathbf{f}_S$ the set of feature vectors corresponding to labelled inputs and by $\mathbf{f}_Q$ the set of feature vectors corresponding to unlabelled inputs. For a feature vector $\mathbf{f}\in \mathbf{f}_S\cup\mathbf{f}_Q$, we denote by $\ell(\mathbf{f})$ the corresponding label. We use $0 < i \leq wq$ to denote the index of an unlabelled sample, so that $\mathbf{f}_Q = (\mathbf{f}_i)_i$, and we denote $\mathbf{c}_j, 0 < j \leq w$ the estimated center for feature vectors corresponding to class $j$.

Our algorithm consists in several steps in which we estimate class centers from a soft allocation matrix $\mathbf{M^*}$, then we update the allocation matrix based on the newly found class centers and iterate the process. In the following paragraphs, we detail these steps.

\paragraph{Sinkhorn mapping.} Considering using MAP estimation for the class centers, and assuming a Gaussian distribution for each class, we typically aim at solving:
\begin{equation}
\resizebox{0.9\hsize}{!}{$
\begin{array}{rcl}
    \{\hat{l}(\mathbf{f}_i) \}, \{\hat{\mathbf{c}}_j\}& = &  \arg\max _{\{\ell(\mathbf{f}_i)\} \in \mathcal{C}, \{\mathbf{c}_j \}} \prod_i P(\mathbf{f}_i | j=\ell(\mathbf{f}_i))  \\
     & =& \arg\min_{\{\ell(\mathbf{f}_i)\} \in \mathcal{C}, \{\mathbf{c}_j \}} \sum_i (\mathbf{f}_i - \mathbf{c}_{\ell(\mathbf{f}_i)})^2,
\end{array}
$}
\label{eq:MAP}
\end{equation}
where $\mathcal{C}$ represents the set of admissible labelling sets. Let us point out that the last term corresponds exactly to the Wasserstein distance used in the Optimal Transport problem formulation~\cite{cuturi2013sinkhorn}.

Therefore, in this step we find the class mapping matrix that minimizes the Wasserstein distance. Inspired by the Sinkhorn algorithm~\cite{villani2008optimal,cuturi2013sinkhorn}, we define the mapping matrix $\mathbf{M^*}$ as follows:
\begin{equation}
\begin{split}
\mathbf{M^*}&= Sinkhorn(\mathbf{L}, \mathbf{p}, \mathbf{q}, \lambda) \\
&= \arg\min_{\mathbf{M}\in\mathbb{U}(\mathbf{p},\mathbf{q})} \sum_{ij}\mathbf{M}_{ij}\mathbf{L}_{ij}  + \lambda H(\mathbf{M}),
\end{split}
\label{eq:M}
\end{equation}
where $\mathbb{U}(\mathbf{p},\mathbf{q})\in\mathbb{R}_{+}^{wq \times w}$ is a set of positive matrices for which the rows sum to $\mathbf{p}$ and the columns sum to $\mathbf{q}$. Formally, $\mathbb{U}(\mathbf{p},\mathbf{q})$ can be written as:
\begin{equation}
\resizebox{0.9\hsize}{!}{$
\mathbb{U}(\mathbf{p},\mathbf{q}) = \{\mathbf{M}\in\mathbb{R}_{+}^{wq \times w} | \mathbf{M}\mathbf{1}_{w}=\mathbf{p}, \mathbf{M}^T\mathbf{1}_{wq}=\mathbf{q}\},
$}
\label{eq:U}
\end{equation}
$\mathbf{p}$ denotes the distribution of the amount that each unlabelled example uses for class allocation, and $\mathbf{q}$ denotes the distribution of the amount of unlabelled examples allocated to each class. Therefore, $\mathbb{U}(\mathbf{p},\mathbf{q})$ contains all the possible ways of allocating examples to classes. The cost function $\mathbf{L}\in\mathbb{R}^{wq \times w}$ in Equation~(\ref{eq:M}) consists of the euclidean distances between unlabelled examples and class centers, hence $\mathbf{L}_{ij}$ denotes the euclidean distance between example $i$ and class center $j$. Here we assume a soft class mapping, meaning that each example can be ``sliced'' into different classes.

The second term on the right of Equation~(\ref{eq:M}) denotes the entropy of $\mathbf{M}$: $H(\mathbf{M})=-\sum_{ij}\mathbf{M}_{ij}\log \mathbf{M}_{ij}$, regularized by a hyper-parameter $\lambda$. Increasing $\lambda$ would force the entropy to become smaller, so that the mapping is less homogeneous. This term also makes the objective function strictly convex~\cite{cuturi2013sinkhorn, solomon2015convolutional} and thus a practical and effective computation. 
From lemma 2 in~\cite{cuturi2013sinkhorn}, the result of this Sinkhorn mapping has the typical form $\mathbf{M^*} = \text{diag}(\mathbf{u}) \cdot \exp(-\mathbf{L}/\lambda) \cdot \text{diag}(\mathbf{v}).$ 

\paragraph{Iterative center estimation.} In this step, our aim is to estimate class centers.
As shown in Algorithm~\ref{algo}, we initialize $\mathbf{c}_j$ as the average of labelled samples belonging to class $j$. Then $\mathbf{c}_j$ is iteratively re-estimated. At each iteration, we compute a mapping matrix $\mathbf{M^*}$ on the unlabelled examples using the sinkhorn mapping. Along with labelled examples, we re-estimate $\mathbf{c}_j$ (temporarily denoted $\bm{\mu}_j$) by weighted-averaging the features with their allocated portions for class $j$:
\begin{equation}
\bm{\mu}_j = g(\mathbf{M^*},j) =  \frac{\sum_{i=1}^{wq} \mathbf{M^*}_{ij}\mathbf{f}_i+\sum_{\mathbf{f}\in\mathbf{f}_S,\ell(\mathbf{f})=j}\mathbf{f}}{s+\sum_{i=1}^{wq} \mathbf{M^*}_{ij}}.
\label{eq:re-estimation}
\end{equation}

This formula corresponds to the minimization of Equation~(\ref{eq:M}). Note that labelled examples do not participate in the mapping process. Since their labels are known, we instead set allocations for their belonging classes to be $1$ and to the others to be $0$. Therefore, labelled examples have the largest possible weight when re-estimating the class centers. 

\paragraph{Proportioned center update.} In order to avoid taking risky harsh decisions in early iterations of the algorithm, we propose to proportionate the update of class centers using an inertia parameter. In more details, we update the center with a learning rate $0 < \alpha \leq 1$. When $\alpha$ is close to 0, the update becomes very slow, whereas $\alpha = 1$ corresponds to directly allocating the newly found class centers:
\begin{equation}
\mathbf{c}_j \leftarrow \mathbf{c}_j + \alpha (\bm{\mu}_j - \mathbf{c}_j).
\label{eq:update}
\end{equation}

\paragraph{Final decision.} After a fixed number of steps $n_{steps}$, the rows of $\mathbf{M^*}$ are interpreted as probabilities to belong to each class. The maximal value corresponds to the decision of the algorithm.

A summary of our proposed algorithm is presented in Algorithm~\ref{algo}. In Table~\ref{tab:hyperparams} we summarize the main parameters and hyperparameters of the considered problem and proposed solution. The code is available at~\url{XXX}.

\begin{algorithm}[h]
\label{algo}
\SetAlgoLined
\SetKwFor{RepTimes}{repeat}{times:}{end}
\SetKwInOut{Parameter}{Parameters}
\SetKwInOut{Initialization}{Initialization}
\caption{Proposed algorithm}
\Parameter{$w, s, q, \lambda, \alpha, n_{steps}$}
\Initialization{$\mathbf{c}_j= \frac{1}{s}\cdot\sum_{\mathbf{f}\in\mathbf{f}_S,\ell(\mathbf{f})=j}\mathbf{f}$}
\RepTimes{$n_{steps}$}{
    $\mathbf{L}_{ij}=\|\mathbf{f}_i-\mathbf{c}_j\|^2, \forall i,j$\\
    $\mathbf{M^*}=Sinkhorn(\mathbf{L}, \mathbf{p}=\mathbf{1}_{wq}, \mathbf{q}=q\mathbf{1}_{w}, \lambda)$\\
    $\bm{\mu}_j = g(\mathbf{M^*},j)$\\ 
    $\mathbf{c}_j \leftarrow \mathbf{c}_j + \alpha (\bm{\mu}_j - \mathbf{c}_j)$\\
}
\Return $\hat{\ell}(\mathbf{f}_i)=\arg\max_j(\mathbf{M^*}[i,j])$
\end{algorithm}

\begin{table*}[h]
    \caption{Important parameters and hyperparameters.}
    \centering
    \scalebox{0.8}{
    \begin{tabular}{|c|c|c|}
    \hline
    \multicolumn{3}{|c|}{\textbf{Novel dataset parameters}}\\
    \hline
    Notation & Value & Description\\
    \hline
         $w$& typically 5 & number of classes\\
         \hline
         $s$& typically 1 or 5 &number of labelled inputs per class\\
         \hline
         $q$ & typically 15 & number of unlabelled inputs per class\\
         \hline
         \hline

    \multicolumn{3}{|c|}{\textbf{Proposed method hyperparameters}}\\
\hline
Notation & Range & Description \\
\hline
      $\beta$ &   $\{-2,-1,-0.5,0,0.5,1,2\}$ & coefficient to adjust distribution skew\\
         \hline
        $\lambda$ & $\lambda\in\mathbb{R}_{+}$ & regularization coefficient for sinkhorn mapping\\
         \hline
         $\alpha$ & $0< \alpha \leq 1$ & learning rate for class center updates\\
         \hline
    \end{tabular}
    }
    \label{tab:hyperparams}
\end{table*}

\section{Experiments}
\label{experiments}
\subsection{Datasets}

We evaluate the performance of the proposed method using standardized few-shot classification datasets: miniImageNet~\cite{vinyals2016matching}, tieredImageNet~\cite{ren2018meta}, CUB~\cite{wah2011caltech} and CIFAR-FS~\cite{bertinetto2018meta}. The \textbf{miniImageNet} dataset contains 100 classes randomly chosen from ILSVRC-
2012~\cite{russakovsky2015imagenet} and 600 images of size $84\times84$ pixels per class. It is split into 64 base classes, 16 validation classes and 20 novel classes. The \textbf{tieredImageNet} dataset is another subset of ImageNet, it consists of 34 high-level categories with 608 classes in total. These categories are split into 20 meta-training superclasses, 6 meta-validation superclasses and 8 meta-test superclasses, which corresponds to 351 base classes, 97 validation classes and 160 novel classes respectively. The \textbf{CUB} dataset contains 200 classes and has 11,788 images of size $84\times84$ pixels in total. Following~\cite{hu2020exploiting}, it is split into 100 base classes, 50 validation classes and 50 novel classes. The \textbf{CIFAR-FS} dataset has 100 classes, each class contains 600 images of size $32\times32$ pixels. The splits of this dataset are the same as those in miniImageNet.

\subsection{Implementation details}
In order to stress the genericity of our proposed method with regards to the chosen backbone architecture and training strategy, we perform experiments using \textbf{WRN}~\cite{zagoruyko2016wide}, \textbf{ResNet18} and \textbf{ResNet12}~\cite{he2016deep}, along with some other pretrained backbones (e.g. DenseNet~\cite{huang2017densely}). For each dataset we train the feature extractor with base classes, tune the hyperparameters with validation classes and test the performance using novel classes. Therefore, for each test run, $w$ classes are drawn uniformly at random among novel classes. Among these $w$ classes, $s$ labelled examples and $q$ unlabelled examples per class are uniformly drawn at random to form $\mathbf{D}_{novel}$. The WRN and ResNet are trained following~\cite{mangla2020charting}. In the inductive setting, we use our proposed Power Transform followed by a basic Nearest Class Mean (NCM) classifier. In the transductive setting, the MAP or an alternative is applied after PT. In order to better segregate between feature vectors of corresponding classes for each task, we implement the ``trans-mean-sub''~\cite{lichtenstein2020tafssl} before MAP where we separately subtract inputs by the means of labelled and unlabelled examples, followed by a unit hypersphere projection. All our experiments are performed using $w=5, q=15$, $s=1$ or $5$. We run 10,000 random draws to obtain mean accuracy score and indicate confidence scores ($95\%$) when relevant. The tuned hyperparameters for miniImageNet are $\beta=0.5, \lambda=10, \alpha=0.4$ and $n_{steps}=30$ for $s=1$; $\beta=0.5, \lambda=10, \alpha=0.2$ and $n_{steps}=20$ for $s=5$. Hyperparameters for other datasets are detailed in the experiments below. 

\subsection{Comparison with state-of-the-art methods}

In the first experiment, we conduct our proposed method on different benchmarks and compare the performance with other state-of-the-art solutions. The results are presented in Table~\ref{tab:results}, we observe that our method with WRN as backbone reaches the state-of-the-art performance for most cases in both inductive and transductive settings on all the benchmarks. In Table~\ref{tab:tiered_results} we also implement our proposed method on tieredImageNet based on a pre-trained DenseNet121 backbone following the procedure described in~\cite{wang2019simpleshot}. From these experiments we conclude that the proposed method can bring an increase of accuracy with a variety of backbones and datasets, leading to competitive performance. In terms of execution time, we measured an average of $0.002s$ per run.

\begin{table*}[h]
    \caption{1-shot and 5-shot accuracy of state-of-the-art methods in the literature, compared with the proposed solution. We present results using WRN as backbones for our proposed solutions.}
    \centering
    \scalebox{0.8}{
    \begin{tabular}{c|l|l|l|l}
         \toprule
         &     &     & \multicolumn{2}{c}{\textbf{miniImageNet}} \\
         Setting & Method & Backbone & 1-shot & 5-shot \\
         \midrule
         \multirow{7}{*}{Inductive}
         &Baseline++~\cite{chen2019closer} & ResNet18 & $51.87\pm0.77\%$ & $75.68\pm0.63\%$\\
         &MAML~\cite{finn2017model} & ResNet18 & $49.61\pm0.92\%$ & $65.72\pm0.77\%$\\
         &ProtoNet~\cite{snell2017prototypical} & WRN & $62.60\pm0.20\%$ & $79.97\pm0.14\%$\\
         &Matching Networks~\cite{vinyals2016matching} & WRN & $64.03\pm0.20\%$ & $76.32\pm0.16\%$\\ 
         &SimpleShot~\cite{wang2019simpleshot} & DenseNet121 & $64.29\pm0.20\%$ & $81.50\pm0.14\%$\\
         &S2M2\_R~\cite{mangla2020charting} & WRN & $64.93\pm0.18\%$ & $83.18\pm0.11\%$\\
         &PT+NCM(ours) & WRN & $\mathbf{65.35\pm0.20}\%$ & $\mathbf{83.87\pm0.13}\%$\\
         \midrule
         \multirow{5}{*}{Transductive}
         &BD-CSPN~\cite{liu2019prototype} & WRN & $70.31\pm0.93\%$ & $81.89\pm0.60\%$\\
         &Transfer+SGC~\cite{hu2020exploiting} & WRN & $76.47\pm0.23\%$ & $85.23\pm0.13\%$\\
         &TAFSSL~\cite{lichtenstein2020tafssl} & DenseNet121 & $77.06\pm0.26\%$ & $84.99\pm0.14\%$\\
         &DFMN-MCT~\cite{kye2020transductive} & ResNet12 & $78.55\pm0.86\%$ & $86.03\pm0.42\%$\\
         &PT+MAP(ours) & WRN & $\mathbf{82.92\pm0.26}\%$ & $\mathbf{88.82\pm0.13}\%$\\ 
         \bottomrule
         
         \toprule
         &     &      & \multicolumn{2}{c}{\textbf{CUB}} \\
         Setting & Method & Backbone & 1-shot & 5-shot \\       
         \midrule
         \multirow{6}{*}{Inductive}
         &Baseline++~\cite{chen2019closer} & ResNet10 & $69.55\pm0.89\%$ & $85.17\pm0.50\%$\\
         &MAML~\cite{finn2017model} & ResNet10 & $70.32\pm0.99\%$ & $80.93\pm0.71\%$\\
         &ProtoNet~\cite{snell2017prototypical} & ResNet18 & $72.99\pm0.88\%$ & $86.64\pm0.51\%$\\
         &Matching Networks~\cite{vinyals2016matching} & ResNet18 & $73.49\pm0.89\%$ & $84.45\pm0.58\%$\\ 
         &S2M2\_R~\cite{mangla2020charting} & WRN & $\mathbf{80.68\pm0.81}\%$ & $90.85\pm0.44\%$\\
         &PT+NCM(ours) & WRN & $80.57\pm0.20\%$ & $\mathbf{91.15\pm0.10}\%$\\
         \midrule
         \multirow{3}{*}{Transductive}
         &BD-CSPN~\cite{liu2019prototype} & WRN & $87.45\%$ & $91.74\%$\\
         &Transfer+SGC~\cite{hu2020exploiting} & WRN & $88.35\pm0.19\%$ & $92.14\pm0.10\%$\\
         &PT+MAP(ours) & WRN & $\mathbf{91.55\pm0.19}\%$ & $\mathbf{93.99\pm0.10}\%$\\ 
         \bottomrule
         
         \toprule
         &     &     & \multicolumn{2}{c}{\textbf{CIFAR-FS}} \\
         Setting & Method & Backbone & 1-shot & 5-shot \\
         \midrule
         \multirow{4}{*}{Inductive}
         &ProtoNet~\cite{snell2017prototypical} & ConvNet64 & $55.50\pm0.70\%$ & $72.00\pm0.60\%$\\
         &MAML~\cite{finn2017model} & ConvNet32 & $58.90\pm1.90\%$ & $71.50\pm1.00\%$\\
         &S2M2\_R~\cite{mangla2020charting} & WRN & $\mathbf{74.81\pm0.19}\%$ & $87.47\pm0.13\%$\\
         &PT+NCM(ours) & WRN & $74.64\pm0.21\%$ & $\mathbf{87.64\pm0.15}\%$\\
         \midrule
         \multirow{3}{*}{Transductive}
         &DSN-MR~\cite{simon2020adaptive} & ResNet12 & $78.00\pm0.90\%$ & $87.30\pm0.60\%$\\
         &Transfer+SGC~\cite{hu2020exploiting} & WRN & $83.90\pm0.22\%$ & $88.76\pm0.15\%$\\
         &PT+MAP(ours) & WRN & $\mathbf{87.69\pm0.23\%}$ & $\mathbf{90.68\pm0.15}\%$\\
         \bottomrule
    \end{tabular}
    }
    \label{tab:results}
\end{table*}

\paragraph{Performance on cross-domain settings.} We also test our method in a cross-domain setting, where the backbone is trained with the base classes in miniImageNet but tested with the novel classes in CUB dataset. As shown in Table~\ref{tab:results_cross}, the proposed method gives the best accuracy both in the case of 1-shot and 5-shot. 

\begin{table}[h]
    \caption{1-shot and 5-shot accuracy of state-of-the-art methods on tieredImageNet.}
    \centering
    \scalebox{0.7}{
    \begin{tabular}{l|l|l|l}
         \toprule
         &          & \multicolumn{2}{c}{\textbf{tieredImageNet}} \\
         Method & Backbone & 1-shot & 5-shot \\       
         \midrule
         ProtoNet~\cite{snell2017prototypical}$^{\flat}$ & ConvNet4 & $53.31\pm0.89\%$ & $72.69\pm0.74\%$\\
         LEO~\cite{rusu2018meta}$^{\flat}$ & WRN & $66.33\pm0.05\%$ & $81.44\pm0.09\%$\\
         SimpleShot~\cite{wang2019simpleshot}$^{\flat}$ & DenseNet121 & $\mathbf{71.32\pm0.22\%}$ & $\mathbf{86.66\pm0.15\%}$\\
         PT+NCM(ours)$^{\flat}$ & DenseNet121 & $69.96\pm0.22\%$ & $86.45\pm0.15\%$\\ 
         \midrule
         DFMN-MCT~\cite{kye2020transductive}$^{\sharp}$ & ResNet12 & $80.89\pm0.84\%$ & $87.30\pm0.49\%$\\
         TAFSSL~\cite{lichtenstein2020tafssl}$^{\sharp}$ & DenseNet121 & $84.29\pm0.25\%$ & $89.31\pm0.15\%$\\
         PT+MAP(ours)$^{\sharp}$ & DenseNet121 & $\mathbf{85.67\pm0.26}\%$ & $\mathbf{90.45\pm0.14}\%$\\ 
         \bottomrule
         \multicolumn{4}{l}{%
            \begin{minipage}{4.5cm}%
            \small $^{\flat}$: Inductive setting.\\
            $^{\sharp}$: Transductive setting. %
            \end{minipage}%
        }
    \end{tabular}
    }
    \label{tab:tiered_results}
\end{table}

\begin{table}[h]
    \caption{1-shot and 5-shot accuracy of state-of-the-art methods when performing cross-domain classification (backbone: WRN).}
    \centering
    \scalebox{0.8}{
    \begin{tabular}{l|l|l}
         \toprule
         Method & 1-shot & 5-shot \\
         \midrule
         Baseline++~\cite{chen2019closer}$^{\flat}$ & $40.44\pm0.75\%$ & $56.64\pm0.72\%$\\
         Manifold Mixup~\cite{verma2018manifold}$^{\flat}$ & $46.21\pm0.77\%$ & $66.03\pm0.71\%$\\
         S2M2\_R~\cite{mangla2020charting}$^{\flat}$ & $48.24\pm0.84\%$ & $\mathbf{70.44\pm0.75}\%$\\
         PT+NCM(ours)$^{\flat}$  & $\mathbf{48.37\pm0.19}\%$ & $70.22\pm0.17\%$\\
         \midrule
         Transfer+SGC~\cite{hu2020exploiting}$^{\sharp}$ & $58.63\pm0.25\%$ & $73.46\pm0.17\%$\\
         PT+MAP(ours)$^{\sharp}$  & $\mathbf{62.49\pm0.32}\%$ & $\mathbf{76.51\pm0.18}\%$\\
         \bottomrule
         \multicolumn{3}{l}{%
            \begin{minipage}{4.5cm}%
            \small $^{\flat}$: Inductive setting.\\
            $^{\sharp}$: Transductive setting. %
            \end{minipage}%
        }
    \end{tabular}
    }
    \label{tab:results_cross}
\end{table}

\subsection{Other experiments}

\paragraph{Ablation study.} To prove the interest of the ingredients on the proposed method in order to reach top performance, we report in Tables~\ref{tab:results_sup} and~\ref{tab:results_compare} the results of ablation studies. In Table~\ref{tab:results_sup}, we first investigate the impact of changing the backbone architecture. Together with previous experiments, we observe that the proposed method consistently achieves the best results for any fixed backbone architecture. We also report performance in the case of inductive few-shot using a simple Nearest-Class Mean (NCM) classifier instead of the iterative MAP procedure described in Section~\ref{methodology}.
We perform another experiment where we replace the MAP algorithm with a standard K-Means where centroids are initialized with the available labelled samples for each class. We can observe significant drops in accuracy, emphasizing the interest of the proposed MAP procedure to better estimate the class centers.

In Table~\ref{tab:results_compare} we show the impact of PT in the transductive setting, where we can see about $6\%$ gain for 1-shot and $4\%$ gain for 5-shot in terms of accuracy.

\begin{table*}[h]
    \caption{Accuracy of the proposed method in inductive and transductive settings, with different backbones, and comparison with K-Means and NCM baselines.}
    \centering
    \scalebox{0.7}{
    \begin{tabular}{l|l|l|l|l|l|l|l}
         \toprule
              \multicolumn{2}{c|}{\textbf{Setting}}   & \multicolumn{2}{c|}{\textbf{Inductive}} & \multicolumn{4}{c}{\textbf{Transductive}} \\
         \midrule
              &          & \multicolumn{2}{c|}{\textbf{(NCM baseline) Proposed PT+NCM}} & \multicolumn{2}{c|}{\textbf{PT+K-Means}} & \multicolumn{2}{c}{\textbf{Proposed PT+MAP}}\\
         Dataset & Backbone & 1-shot & 5-shot & 1-shot & 5-shot & 1-shot & 5-shot\\
         \midrule
         \multirow{3}{*}{miniImageNet} & ResNet12 & ($49.08$) $62.68\pm0.20\%$ & ($70.85$) $81.99\pm0.14\%$ & $72.73\pm0.23\%$& $84.05\pm0.14\%$ & $78.47\pm0.28\%$ & $85.84\pm0.15\%$\\
         & ResNet18 & ($47.63$) $62.50\pm0.20\%$ & ($72.89$) $82.17\pm0.14\%$& $73.08\pm0.22\%$ & $84.67\pm0.14\%$ &$80.00\pm0.27\%$ & $86.96\pm0.14\%$\\
         & WRN & ($55.31$) $\mathbf{65.35\pm0.20}\%$ & ($78.33$) $\mathbf{83.87\pm0.13}\%$ & $\mathbf{76.67\pm0.22}\%$ & $\mathbf{86.73\pm0.13}\%$ & $\mathbf{82.92\pm0.26}\%$ & $\mathbf{88.82\pm0.13}\%$ \\
         \midrule
         \multirow{3}{*}{CUB} & ResNet12 & ($61.30$) $78.40\pm0.20\%$ & ($82.83$) $91.12\pm0.10\%$& $87.35\pm0.19\%$ & $92.31\pm0.10\%$ & $90.96\pm0.20\%$ & $93.77\pm0.09\%$\\
         & ResNet18 & ($58.92$) $76.98\pm0.20\%$ & ($82.69$) $90.56\pm0.10\%$& $87.16\pm0.19\%$ & $91.97\pm0.09\%$ &  $91.10\pm0.20\%$ & $93.78\pm0.09\%$\\
         & WRN & ($69.21$) $\mathbf{80.57\pm0.20}\%$ & ($88.33$) $\mathbf{91.15\pm0.10}\%$&$\mathbf{88.28\pm0.19}\%$ & $\mathbf{92.37\pm0.10}\%$ & $\mathbf{91.55\pm0.19}\%$ & $\mathbf{93.99\pm0.10}\%$ \\
         \midrule
         \multirow{3}{*}{CIFAR-FS} & ResNet12 &($52.50$) $71.02\pm0.22\%$ & ($74.16$) $84.68\pm0.16\%$ & $78.39\pm0.24\%$& $85.73\pm0.16\%$ & $82.45\pm0.27\%$ & $87.33\pm0.17\%$\\
         & ResNet18 &($56.40$) $71.41\pm0.22\%$ &($78.30$) $85.50\pm0.15\%$ &$79.95\pm0.23\%$ & $86.74\pm0.16\%$ &  $84.80\pm0.25\%$ & $88.55\pm0.16\%$\\
         & WRN &($68.93$) $\mathbf{74.64\pm0.21}\%$ &($86.81$) $\mathbf{87.64\pm0.15}\%$ &$\mathbf{83.69\pm0.22}\%$ & $\mathbf{89.19\pm0.15}\%$ & $\mathbf{87.69\pm0.23\%}$ & $\mathbf{90.68\pm0.15}\%$ \\
         \bottomrule
         
    \end{tabular}
    }
    \label{tab:results_sup}
\end{table*}

\begin{table*}[h]
    \caption{Influence of Power Transform in the transductive setting with different backbones on miniImageNet.}
    \centering
    \scalebox{0.8}{
    \begin{tabular}{p{0.5cm}|p{1.5cm}|c|c|c|c|c|c}
         \toprule
         &  & \multicolumn{2}{c}{\textbf{WRN}} & \multicolumn{2}{c}{\textbf{ResNet18}}& \multicolumn{2}{c}{\textbf{ResNet12}}\\
         PT  & MAP & 1-shot & 5-shot & 1-shot & 5-shot & 1-shot & 5-shot\\
         \midrule
         & \checkmark &$75.60\pm0.29\%$ & $84.13\pm0.16\%$ & $74.48\pm0.29\%$ & $82.88\pm0.17\%$ &$72.04\pm0.30\%$ & $80.98\pm0.18\%$\\
         
         \midrule
         \checkmark &\checkmark & $\mathbf{82.92\pm0.26}\%$ & $\mathbf{88.82\pm0.13}\%$& $\mathbf{80.00\pm0.27}\%$ & $\mathbf{86.96\pm0.14}\%$ &$\mathbf{78.47\pm0.28}\%$ &$\mathbf{85.84\pm0.15}\%$\\
         
         \bottomrule
         
    \end{tabular}
    }
    \label{tab:results_compare}
\end{table*}

\begin{figure}[h]
\begin{center}
\scalebox{0.7}{
    \begin{tikzpicture}
    \begin{axis}[
        axis lines = left,
    ]
    \addplot [
        domain=0:0.5, 
        samples=50, 
        color=red,
        thick,
    ]
        {31.88941138 * exp(-3.54486796 * x) - 2.81381571};
    \addlegendentry{Class 1}
    \addplot [
        domain=0:0.5, 
        samples=50, 
        color=blue,
        thick,
        ]
        {132.31347387 * exp(-26.48181533 * x) + 0.36254811};
    \addlegendentry{Class 2}
    
    \addplot [
        domain=0:0.5, 
        samples=50, 
        color=green,
        thick,
        ]
        {58.71077998 * exp(-10.0106652 * x) - 0.10442753};
    \addlegendentry{Class 3}
    
    \addplot [
        domain=0:0.5, 
        samples=50, 
        color=orange,
        thick,
        ]
        {70.49026213 * exp(-11.88730944 * x) - 0.23004155};
    \addlegendentry{Class 4}
    
    \addplot [
        domain=0:0.5, 
        samples=50, 
        color=cyan,
        thick,
        ]
        {192.96327014 * exp(-45.39826927 * x) + 0.75375176};
    \addlegendentry{Class 5}
    \end{axis}
    \node [anchor=center] at (rel axis cs: 0.5,-0.18) {$(a)$};
    \begin{axis}[
        yshift=-75mm,
        axis lines = left,
    ]
    \addplot [
        domain=0:0.5, 
        samples=50, 
        color=red,
        thick,
        ]
        {8.22628596e+01 * exp(-(x - 7.22708764e-02)^2 / (2 * 2.97571297e-02^2))};
    \addlegendentry{Class 1}
    \addplot [
        domain=0:0.5, 
        samples=50, 
        color=blue,
        thick,
        ]
        {5.80822496e+01 * exp(-(x - 5.56072354e-02)^2 / (2 * 4.48482135e-02^2))};
    \addlegendentry{Class 2}
    
    \addplot [
        domain=0:0.5, 
        samples=50, 
        color=green,
        thick,
        ]
        {6.55055182e+01 * exp(-(x - 6.03167579e-02)^2 / (2 * 3.73229630e-02^2))};
    \addlegendentry{Class 3}
    
    \addplot [
        domain=0:0.5, 
        samples=50, 
        color=orange,
        thick,
        ]
        {6.29905894e+01 * exp(-(x - 6.49222465e-02)^2 / (2 * 3.96683481e-02^2))};
    \addlegendentry{Class 4}
    
    \addplot [
        domain=0:0.5, 
        samples=50, 
        color=cyan,
        thick,
        ]
        {5.47471994e+01 * exp(-(x - 4.67235457e-02)^2 / (2 * 5.11186492e-02^2))};
    \addlegendentry{Class 5}
    \end{axis}
    \node [anchor=center] at (rel axis cs: 0.5,-1.5) {$(b)$};
    \end{tikzpicture}
    
}
\end{center}
\caption{Distributions of an arbitrarily chosen feature for 5 novel classes before $(a)$ and after $(b)$ PT.}
\label{fig:analyse}
\end{figure}
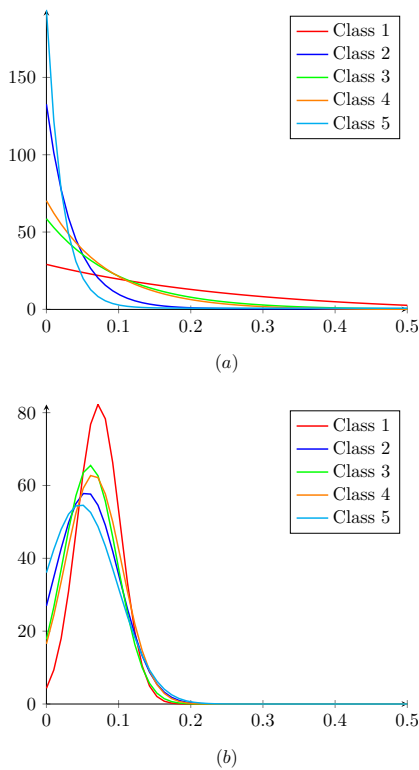

\paragraph{Effect of Power Transform.} To visualize the effect of PT on the feature distributions, we depict in Figure~\ref{fig:analyse} the distributions of an arbitrarily selected feature for 5 randomly selected novel classes of miniImageNet when using WRN, before and after applying PT. We observe quite clearly how PT is able to reshape the feature distributions to close-to-gaussian distributions. We observed similar behaviors with other datasets as well.

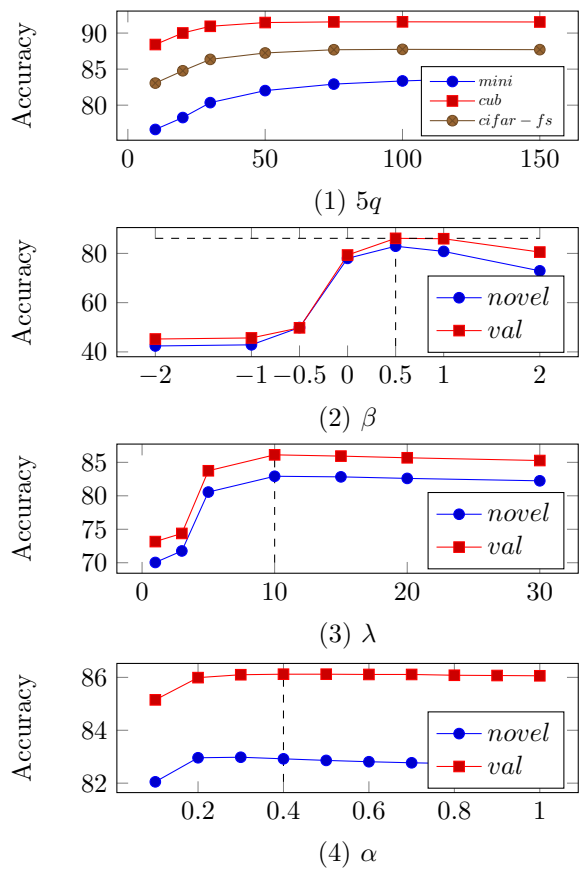
\begin{figure}[h!]
  \begin{center}
    \begin{tikzpicture}
       \begin{scope}[]
        \begin{axis}[
            xlabel=(1) $5q$,
            ylabel=Accuracy,
            height=3.3cm,
            width=.45\textwidth,
            legend style={nodes={scale=0.6, transform shape}},
            legend pos={south east,legend cell align=left},
            ]
          
          \addlegendentry{$mini$}
          \addplot coordinates
          {(10,76.63) (20,78.28) (30,80.35) (50,82.02) (75,82.92) (100,83.36) (150,83.95)};
          
          
          \addlegendentry{$cub$}
          \addplot coordinates
          {(10,88.41) (20,90.01) (30,90.94) (50,91.46) (75,91.55) (100,91.56) (150,91.54)};
          
          \addlegendentry{$cifar-fs$}
          \addplot coordinates
          {(10,83.06) (20,84.77) (30,86.35) (50,87.24) (75,87.69) (100,87.74) (150,87.70)};
          
        \end{axis}
      \end{scope}
    \end{tikzpicture}
    \begin{tikzpicture}
       \begin{scope}[]
        \begin{axis}[
            xlabel=(2) $\beta$,
            ylabel=Accuracy,
            height=3.3cm,
            width=.45\textwidth,
            ytick = {40, 60, 80},
            xtick = {-2.5, -2, -1, -0.5, 0, 0.5, 1, 2, 2.5},
            legend pos={south east,legend cell align=left}
            ]
          
          \addlegendentry{$novel$}
          \addplot coordinates
          {(-2,42.41) (-1,42.87) (-0.5,49.89) (0,77.98) (0.5,82.92) (1,80.81) (2,72.92)};
          
          \addlegendentry{$val$}
          \addplot coordinates
          {(-2,45.23) (-1,45.67) (-0.5,49.78) (0,79.34) (0.5,86.12) (1,85.96) (2,80.54)};
          
          \addplot [dashed, mark options={scale=1}] coordinates {(0.5,42.41) (0.5,86.12)};
          
          \addplot [dashed, mark options={scale=1}] coordinates {(-2,86.12) (2,86.12)};
        \end{axis}
      \end{scope}
    \end{tikzpicture}
    \begin{tikzpicture}
       \begin{scope}[]
        \begin{axis}[
            xlabel=(3) $\lambda$,
            ylabel=Accuracy,
            height=3.3cm,
            width=.45\textwidth,
            legend pos={south east,legend cell align=left}
            ]
          
          \addlegendentry{$novel$}
          \addplot coordinates
          {(1,70.04) (3,71.76) (5,80.56) (10,82.92) (15,82.83) (20,82.60) (30,82.24)};
          
          \addlegendentry{$val$}
          \addplot coordinates
          {(1,73.16) (3,74.38) (5,83.74) (10,86.12) (15,85.93) (20,85.68) (30,85.27)};
          
          \addplot [dashed, mark options={scale=1}] coordinates {(10,70.04) (10,86.12)};
        \end{axis}
      \end{scope}
    \end{tikzpicture}
    \begin{tikzpicture}
       \begin{scope}[]
        \begin{axis}[
            xlabel=(4) $\alpha$,
            ylabel=Accuracy,
            height=3.3cm,
            width=.45\textwidth,
            legend pos={south east,legend cell align=left}
            ]
          
          \addlegendentry{$novel$}
          \addplot coordinates
          {(0.1,82.05) (0.2,82.96) (0.3,82.98) (0.4,82.92) (0.5,82.86) (0.6,82.81) (0.7,82.77) (0.8,82.74) (0.9,82.72) (1,82.68)};
          
          \addlegendentry{$val$}
          \addplot coordinates
          {(0.1,85.15) (0.2,85.99) (0.3,86.10) (0.4,86.12) (0.5,86.12) (0.6,86.11) (0.7,86.11) (0.8,86.08) (0.9,86.07) (1,86.06)};
          
          \addplot [dashed, mark options={scale=1}] coordinates {(0.4,82.05) (0.4,86.12)};
        \end{axis}
      \end{scope}
    \end{tikzpicture}
  \end{center}
  \vspace{-.5cm}
  \caption{(1) represents 5-way 1-shot accuracy on miniImagenet, CUB and CIFAR-FS (backbone: WRN) as a function of $q$. (2), (3) and (4) represent 1-shot accuracy on miniImageNet (backbone: WRN) as a function of $\beta,\lambda$ and $\alpha$ respectively.}
  \label{fig:functionofpara}
\end{figure}

\paragraph{Influence of the number of unlabelled samples.} Small values of $q$ lead to settings that are closer to the inductive case. In order to better understand the gain in accuracy due to having access to more unlabelled samples, we depict in Figure~\ref{fig:functionofpara} the evolution of accuracy as a function of $q$, when $w=5$ is fixed. Interestingly, the accuracy quickly reaches a close-to-asymptotical plateau, emphasizing the ability of the method to soon exploit available information in the task.

\begin{figure}[h]
  \begin{center}
    \begin{tikzpicture}
       \begin{scope}[]
        \begin{axis}[
            xlabel={$q_1$},
            ylabel=Accuracy,
            height=5cm,
            width=.45\textwidth,
            legend style={nodes={scale=0.75, transform shape}, at={(0.0,0.35)}, anchor=west},
            ]
          
          \addlegendentry{\tiny{PT+NCM}}
          \addplot coordinates
          {(1,83.94) (10,83.80) (20,83.70) (30,83.96) (40,83.86) (50,83.96)};
          
          
          \addlegendentry{\tiny{PT+MAP ($\mathbf{q}=[30, 30]$)}}
          \addplot coordinates
          {(30,92.23) (40,94.75) (50,95.49)};
          
          \addlegendentry{\tiny{PT+MAP ($\mathbf{q}=[q_1, 100-q_1]$)}}
          \addplot coordinates
          {(1,99.07) (10,97.02) (20,96.26) (30,95.98) (40,95.92) (50,95.90)};
          
        \end{axis}
      \end{scope}
    \end{tikzpicture}
  \end{center}
  \vspace{-.5cm}
  \caption{Accuracy of 2-ways classification on miniImagenet (1-shot) with unevenly distributed query data for each class in different settings, where the total number of query inputs remains constant (total: 100 elements). When $q_1=1$, we obtain the most imbalanced case, whereas $q_1=50$ corresponds to a balanced case. }
  \label{fig:imbalanced}
\end{figure}
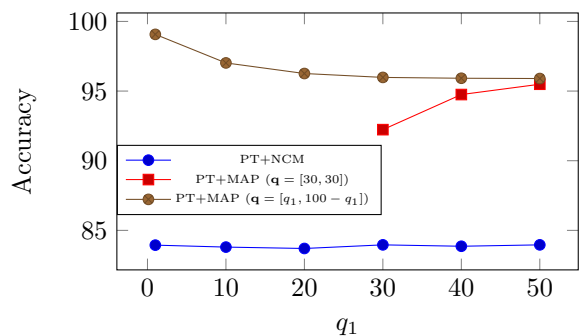

\paragraph{Impact of class imbalance.} In all previous transductive experiments, we assumed a balanced number of unlabelled samples per class. We now consider the case of 2 classes, where we vary the number of unlabelled examples $q1$ of class $1$ with respect to that of class 2 ($100-q1$). In Figure~\ref{fig:imbalanced} we depict: 1) the performance of the inductive version of our method (PT-NCM), which is independent of $q1$, 2) the performance of the proposed transductive method when the vector $\mathbf{q}$ is appropriately defined (knowing the proportion of elements in class 1 vs. class 2), and 3) a mixed case where we expect at least 30 elements in both classes but do not know exactly how many ($\mathbf{q} = [30,30]$). Interestingly, we observe that the transductive setting still outperforms the inductive ones even when the proportion of elements in both classes is only approximately known.

\paragraph{Hyperparameter tuning.} In the next experiment we tune $\beta,\lambda$ and $\alpha$ on the validation classes of each dataset, and then apply them to test our model on novel classes. We vary each hyperparamter in a certain range and observe the evolution of accuracy to choose the peak that corresponds to the highest prediction. For example, the evolving curve for $\beta, \lambda$ and $\alpha$ with miniImageNet are presented in Figure~\ref{fig:functionofpara} (2) to (4). For comparison purposes, we also trace the corresponding curves on novel classes. We draw a dash line on the hyperparameter values where the accuracy on the validation classes peaks, meaning that this is the chosen value resulting in Table~\ref{tab:results}.

The following observations can be drawn from this experiment: 1) The evolving curves on validation classes (red) and novel classes (blue) have generally similar trend for each hyperparameter. In particular, two curves peak at the same $\beta$ $(\beta=0.5)$ and $\lambda$ $(\lambda=10)$, meaning that validation classes and novel classes share the same $\beta$ and $\lambda$ that reach the highest accuracy. 2) A small $\lambda$ tends to lead to a homogeneous class partition for $\mathbf{M}^*$, where each sample are uniformly allocated to $w$ classes. Hence the sharp drop on the accuracy when $\lambda<5$. 3) A too small $\alpha$ results in an insufficient class center update. On the contrary, the impact on a large $\alpha$ is relatively mild. Overall, it is interesting to point out the little sensitivity of the proposed method accuracy with regards to hyperparameter tuning.

We followed this procedure to find the tuned hyperparameters for each dataset. Therefore, we obtained that working with CUB leads to the the same hyperparameters as miniImageNet. For tieredImageNet and CIFAR-FS, the best accuracy are obtained on validation classes when $\beta=0.5, \lambda=10, \alpha=0.3$ for $s=1$; $\beta=0.5, \lambda=10, \alpha=0.2$ for $s=5$.

\section{Conclusion}
\label{conclusion}

In this paper we introduced a new pipeline to solve the few-shot classification problem. Namely, we proposed to firstly preprocess the raw feature vectors to better align to a Gaussian distribution and then we designed an optimal-transport inspired iterative algorithm to estimate the class centers. Our experimental results on standard vision benchmarks reach state-of-the-art accuracy, with important gains in both 1-shot and 5-shot classification. Moreover, the proposed method can bring gains with a variety of feature extractors, with few hyperparameters. Thus we believe that the proposed method is applicable to many practical problems.

\newpage
\small
\bibliographystyle{abbrv}
\bibliography{AISTATS}

\begin{thebibliography}{10}

\bibitem{bertinetto2018meta}
L.~Bertinetto, J.~F. Henriques, P.~H. Torr, and A.~Vedaldi.
\newblock Meta-learning with differentiable closed-form solvers.
\newblock {\em arXiv preprint arXiv:1805.08136}, 2018.

\bibitem{chapelle2009semi}
O.~Chapelle, B.~Scholkopf, and A.~Zien.
\newblock Semi-supervised learning (chapelle, o. et al., eds.; 2006)[book
  reviews].
\newblock {\em IEEE Transactions on Neural Networks}, 20(3):542--542, 2009.

\bibitem{chen2019closer}
W.-Y. Chen, Y.-C. Liu, Z.~Kira, Y.-C.~F. Wang, and J.-B. Huang.
\newblock A closer look at few-shot classification, 2019.

\bibitem{chen2019image}
Z.~Chen, Y.~Fu, Y.-X. Wang, L.~Ma, W.~Liu, and M.~Hebert.
\newblock Image deformation meta-networks for one-shot learning.
\newblock In {\em Proceedings of the IEEE Conference on Computer Vision and
  Pattern Recognition}, pages 8680--8689, 2019.

\bibitem{cuturi2013sinkhorn}
M.~Cuturi.
\newblock Sinkhorn distances: Lightspeed computation of optimal transport.
\newblock In {\em Advances in neural information processing systems}, pages
  2292--2300, 2013.

\bibitem{das2019two}
D.~Das and C.~G. Lee.
\newblock A two-stage approach to few-shot learning for image recognition.
\newblock {\em IEEE Transactions on Image Processing}, 2019.

\bibitem{dempster1977maximum}
A.~P. Dempster, N.~M. Laird, and D.~B. Rubin.
\newblock Maximum likelihood from incomplete data via the em algorithm.
\newblock {\em Journal of the Royal Statistical Society: Series B
  (Methodological)}, 39(1):1--22, 1977.

\bibitem{finn2017model}
C.~Finn, P.~Abbeel, and S.~Levine.
\newblock Model-agnostic meta-learning for fast adaptation of deep networks.
\newblock In {\em Proceedings of the 34th International Conference on Machine
  Learning-Volume 70}, pages 1126--1135. JMLR. org, 2017.

\bibitem{garcia2017few}
V.~Garcia and J.~Bruna.
\newblock Few-shot learning with graph neural networks.
\newblock {\em arXiv preprint arXiv:1711.04043}, 2017.

\bibitem{gidaris2019generating}
S.~Gidaris and N.~Komodakis.
\newblock Generating classification weights with gnn denoising autoencoders for
  few-shot learning.
\newblock {\em arXiv preprint arXiv:1905.01102}, 2019.

\bibitem{8462273}
V.~{Gripon}, G.~B. {Hacene}, M.~{Löwe}, and F.~{Vermet}.
\newblock Improving accuracy of nonparametric transfer learning via vector
  segmentation.
\newblock In {\em 2018 IEEE International Conference on Acoustics, Speech and
  Signal Processing (ICASSP)}, pages 2966--2970, 2018.

\bibitem{he2016deep}
K.~He, X.~Zhang, S.~Ren, and J.~Sun.
\newblock Deep residual learning for image recognition.
\newblock In {\em Proceedings of the IEEE conference on computer vision and
  pattern recognition}, pages 770--778, 2016.

\bibitem{hu2020exploiting}
Y.~Hu, V.~Gripon, and S.~Pateux.
\newblock Exploiting unsupervised inputs for accurate few-shot classification.
\newblock {\em arXiv preprint arXiv:2001.09849}, 2020.

\bibitem{huang2019few}
G.~Huang, H.~Larochelle, and S.~Lacoste-Julien.
\newblock Are few-shot learning benchmarks too simple?
\newblock {\em arXiv preprint arXiv:1902.08605}, 2019.

\bibitem{huang2017densely}
G.~Huang, Z.~Liu, L.~Van Der~Maaten, and K.~Q. Weinberger.
\newblock Densely connected convolutional networks.
\newblock In {\em Proceedings of the IEEE conference on computer vision and
  pattern recognition}, pages 4700--4708, 2017.

\bibitem{kim2019edge}
J.~Kim, T.~Kim, S.~Kim, and C.~D. Yoo.
\newblock Edge-labeling graph neural network for few-shot learning.
\newblock In {\em Proceedings of the IEEE Conference on Computer Vision and
  Pattern Recognition}, pages 11--20, 2019.

\bibitem{kye2020transductive}
S.~M. Kye, H.~B. Lee, H.~Kim, and S.~J. Hwang.
\newblock Transductive few-shot learning with meta-learned confidence.
\newblock {\em arXiv preprint arXiv:2002.12017}, 2020.

\bibitem{lichtenstein2020tafssl}
M.~Lichtenstein, P.~Sattigeri, R.~Feris, R.~Giryes, and L.~Karlinsky.
\newblock Tafssl: Task-adaptive feature sub-space learning for few-shot
  classification.
\newblock {\em arXiv preprint arXiv:2003.06670}, 2020.

\bibitem{liu2019prototype}
J.~Liu, L.~Song, and Y.~Qin.
\newblock Prototype rectification for few-shot learning.
\newblock {\em arXiv preprint arXiv:1911.10713}, 2019.

\bibitem{liu2018learning}
Y.~Liu, J.~Lee, M.~Park, S.~Kim, E.~Yang, S.~J. Hwang, and Y.~Yang.
\newblock Learning to propagate labels: Transductive propagation network for
  few-shot learning.
\newblock {\em arXiv preprint arXiv:1805.10002}, 2018.

\bibitem{mangla2020charting}
P.~Mangla, N.~Kumari, A.~Sinha, M.~Singh, B.~Krishnamurthy, and V.~N.
  Balasubramanian.
\newblock Charting the right manifold: Manifold mixup for few-shot learning.
\newblock In {\em The IEEE Winter Conference on Applications of Computer
  Vision}, pages 2218--2227, 2020.

\bibitem{mensink2012metric}
T.~Mensink, J.~Verbeek, F.~Perronnin, and G.~Csurka.
\newblock Metric learning for large scale image classification: Generalizing to
  new classes at near-zero cost.
\newblock In {\em European Conference on Computer Vision}, pages 488--501.
  Springer, 2012.

\bibitem{ravi2016optimization}
S.~Ravi and H.~Larochelle.
\newblock Optimization as a model for few-shot learning.
\newblock 2016.

\bibitem{ren2018meta}
M.~Ren, E.~Triantafillou, S.~Ravi, J.~Snell, K.~Swersky, J.~B. Tenenbaum,
  H.~Larochelle, and R.~S. Zemel.
\newblock Meta-learning for semi-supervised few-shot classification.
\newblock {\em arXiv preprint arXiv:1803.00676}, 2018.

\bibitem{russakovsky2015imagenet}
O.~Russakovsky, J.~Deng, H.~Su, J.~Krause, S.~Satheesh, S.~Ma, Z.~Huang,
  A.~Karpathy, A.~Khosla, M.~Bernstein, et~al.
\newblock Imagenet large scale visual recognition challenge.
\newblock {\em International journal of computer vision}, 115(3):211--252,
  2015.

\bibitem{rusu2018meta}
A.~A. Rusu, D.~Rao, J.~Sygnowski, O.~Vinyals, R.~Pascanu, S.~Osindero, and
  R.~Hadsell.
\newblock Meta-learning with latent embedding optimization.
\newblock {\em arXiv preprint arXiv:1807.05960}, 2018.

\bibitem{simon2020adaptive}
C.~Simon, P.~Koniusz, R.~Nock, and M.~Harandi.
\newblock Adaptive subspaces for few-shot learning.
\newblock In {\em Proceedings of the IEEE/CVF Conference on Computer Vision and
  Pattern Recognition}, pages 4136--4145, 2020.

\bibitem{snell2017prototypical}
J.~Snell, K.~Swersky, and R.~Zemel.
\newblock Prototypical networks for few-shot learning.
\newblock In {\em Advances in Neural Information Processing Systems}, pages
  4077--4087, 2017.

\bibitem{solomon2015convolutional}
J.~Solomon, F.~De~Goes, G.~Peyr{\'e}, M.~Cuturi, A.~Butscher, A.~Nguyen, T.~Du,
  and L.~Guibas.
\newblock Convolutional wasserstein distances: Efficient optimal transportation
  on geometric domains.
\newblock {\em ACM Transactions on Graphics (TOG)}, 34(4):1--11, 2015.

\bibitem{thrun2012learning}
S.~Thrun and L.~Pratt.
\newblock {\em Learning to learn}.
\newblock Springer Science \& Business Media, 2012.

\bibitem{torrey2010transfer}
L.~Torrey and J.~Shavlik.
\newblock Transfer learning.
\newblock In {\em Handbook of research on machine learning applications and
  trends: algorithms, methods, and techniques}, pages 242--264. IGI Global,
  2010.

\bibitem{tukey1977exploratory}
J.~W. Tukey.
\newblock {\em Exploratory data analysis}, volume~2.
\newblock Reading, Mass., 1977.

\bibitem{vallender1974calculation}
S.~Vallender.
\newblock Calculation of the wasserstein distance between probability
  distributions on the line.
\newblock {\em Theory of Probability \& Its Applications}, 18(4):784--786,
  1974.

\bibitem{verma2018manifold}
V.~Verma, A.~Lamb, C.~Beckham, A.~Najafi, I.~Mitliagkas, A.~Courville,
  D.~Lopez-Paz, and Y.~Bengio.
\newblock Manifold mixup: Better representations by interpolating hidden
  states.
\newblock {\em arXiv preprint arXiv:1806.05236}, 2018.

\bibitem{villani2008optimal}
C.~Villani.
\newblock {\em Optimal transport: old and new}, volume 338.
\newblock Springer Science \& Business Media, 2008.

\bibitem{vinyals2016matching}
O.~Vinyals, C.~Blundell, T.~Lillicrap, D.~Wierstra, et~al.
\newblock Matching networks for one shot learning.
\newblock In {\em Advances in neural information processing systems}, pages
  3630--3638, 2016.

\bibitem{wah2011caltech}
C.~Wah, S.~Branson, P.~Welinder, P.~Perona, and S.~Belongie.
\newblock The caltech-ucsd birds-200-2011 dataset.
\newblock 2011.

\bibitem{wang2019simpleshot}
Y.~Wang, W.-L. Chao, K.~Q. Weinberger, and L.~van~der Maaten.
\newblock Simpleshot: Revisiting nearest-neighbor classification for few-shot
  learning.
\newblock {\em arXiv preprint arXiv:1911.04623}, 2019.

\bibitem{ye2018learning}
H.-J. Ye, H.~Hu, D.-C. Zhan, and F.~Sha.
\newblock Learning embedding adaptation for few-shot learning.
\newblock {\em arXiv preprint arXiv:1812.03664}, 2018.

\bibitem{zagoruyko2016wide}
S.~Zagoruyko and N.~Komodakis.
\newblock Wide residual networks.
\newblock {\em arXiv preprint arXiv:1605.07146}, 2016.

\bibitem{zhang2019few}
H.~Zhang, J.~Zhang, and P.~Koniusz.
\newblock Few-shot learning via saliency-guided hallucination of samples.
\newblock In {\em Proceedings of the IEEE Conference on Computer Vision and
  Pattern Recognition}, pages 2770--2779, 2019.

\end{thebibliography}

\newpage

\onecolumn
\aistatstitle{Supplementary Materials}
\section{ADDITIONAL EXPERIMENTS}

In this section, we provide additional experiments and results on our proposed method, including a combination of multi-backbones in terms of features. We demonstrate that with PT-MAP, a direct concatenation of different backbone features can increase the performance. 

\subsection{Effect of PT-MAP on pre-trained backones}

In the paper we trained the different backbones following~\cite{mangla2020charting}. To evaluate the generosity of our proposed method, here we tested the performance of PT-MAP based on a set of pre-trained backbones~\cite{wang2019simpleshot} that follow a different training procedure. As in Table~\ref{tab:results_sup1}, we can see that our method is still able to bring a large accuracy increase on all backbones, no matter what their training procedure is. Therefore, this proves the generosity of PT-MAP, which can be applied in various applications.

\begin{table}[h]
    \caption{1-shot and 5-shot accuracy (dataset: miniImagenet) on baseline and our proposed PT-MAP.}
    \centering
    \scalebox{0.8}{
    \begin{tabular}{l|l|l|l|l}
         \toprule
         & \multicolumn{2}{c|}{\textbf{Baseline}} & \multicolumn{2}{c}{\textbf{PT-MAP}}\\
         Backbone & 1-shot & 5-shot & 1-shot & 5-shot \\       
         \midrule
         Conv4 & $33.17\pm0.17\%$ & $63.25\pm0.17\%$ & $58.18\pm0.28\%$ & $70.79\pm0.18\%$ \\
         Mobilenet & $55.70\pm0.20\%$ & $77.46\pm0.15\%$ &$73.58\pm0.29\%$ & $82.81\pm0.15\%$ \\
         ResNet10 & $54.45\pm0.21\%$ & $76.98\pm0.15\%$ &$74.91\pm0.29\%$ & $83.73\pm0.15\%$ \\
         ResNet18 & $56.06\pm0.20\%$ & $78.63\pm0.15\%$ &$77.28\pm0.28\%$ & $85.13\pm0.14\%$\\
         WRN & $57.26\pm0.21\%$ & $78.99\pm0.14\%$ &$78.86\pm0.28\%$ & $86.17\pm0.14\%$\\
         DenseNet121 & $57.81\pm0.21\%$ & $80.43\pm0.15\%$ &$79.98\pm0.28\%$ & $87.19\pm0.13\%$ \\
         \bottomrule
    \end{tabular}
    }
    \label{tab:results_sup1}
\end{table}

\subsection{Effect of PT-MAP on multi-backbones}

To further investigate the effect of our proposed method on the features, we perform a direct concatenation of raw feature vectors extracted from multiple backbones before PT-MAP. In Table~\ref{tab:results_sup2} we chose the feature vectors from three backbones (WRN, ResNet18 and ResNet12) and evaluated the performance with different combinations. We observe that a direct concatenation, depending on the backbones, can bring about $1\%$ gain in both 1-shot and 5-shot settings.

\begin{table}[h]
    \caption{1-shot and 5-shot accuracy (datasets: miniImageNet, CUB and CIFAR-FS) on our proposed PT-MAP with multi-backbones ('+' denotes a concatenation of backbone features).}
    \centering
    \scalebox{0.8}{
    \begin{tabular}{l|l|l|l|l|l|l}
         \toprule
         & \multicolumn{2}{c|}{\textbf{miniImageNet}} & \multicolumn{2}{c|}{\textbf{CUB}} & \multicolumn{2}{c}{\textbf{CIFAR-FS}}\\
         Backbone & 1-shot & 5-shot & 1-shot & 5-shot & 1-shot & 5-shot \\       
         \midrule
         WRN & $82.92\%$ & $88.82\%$ & $91.55\%$ & $93.99\%$ & $87.69\%$ & $90.68\%$ \\
         RN18 & $80.00\%$ & $86.96\%$ & $91.10\%$ & $93.78\%$ & $84.80\%$ & $88.55\%$\\
         RN12 & $78.47\%$ & $85.84\%$ & $90.96\%$ & $93.77\%$ & $82.45\%$ & $87.33\%$ \\
         RN18+RN12 & $81.27\%$ & $87.89\%$ & $93.05\%$ & $95.15\%$ & $86.10\%$ & $89.67\%$\\
         WRN+RN18 & $\mathbf{83.87\%}$ & $\mathbf{89.64\%}$ & $93.28\%$ & $95.27\%$ & $88.05\%$ & $91.18\%$\\
         WRN+RN12 & $83.63\%$ & $89.47\%$ & $93.37\%$ & $95.35\%$ & $87.72\%$ & $90.98\%$\\
         WRN+RN18+RN12 & $83.79\%$ & $89.63\%$ & $\mathbf{94.04}\%$ & $\mathbf{95.76}\%$ & $\mathbf{88.15}\%$ & $\mathbf{91.25}\%$\\
         \bottomrule
    \end{tabular}
    }
    \label{tab:results_sup2}
\end{table}

\vfill

\end{document}